\documentclass{article}
\usepackage{ijcai19}

\usepackage{times}
\usepackage{soul}
\usepackage{url}
\usepackage[hidelinks]{hyperref}
\usepackage[utf8]{inputenc}
\usepackage[small]{caption}
\usepackage{graphicx}
\usepackage{epstopdf}
\usepackage{amsmath}
\usepackage{booktabs}
\usepackage{algorithm}
\usepackage{algorithmic}
\title{GraphNAS: Graph Neural Architecture Search with Reinforcement Learning}

\author{
    Yang Gao $^{1,2}$\and
    Hong Yang $^3$\and
    Peng Zhang $^4$\and
    Chuan Zhou $^{1,2}$\and
    Yue Hu $^{1,2}$
    \affiliations
    $^1$Institute of Information Engineering, Chinese Academy of Sciences, Beijing, China \\
    $^2$School of Cyber Security, University of Chinese Academy of Sciences, Beijing, China\\
    $^3$Centre for Artificial Intelligence, University of Technology Sydney, Australia\\
    $^4$Ant Financial Services Group, Hangzhou, China\\
    \emails
    \{gaoyang, zhouchuan,huyue\}@iie.ac.cn,
    hong.yang@student.uts.edu.au,
    zhangpeng04@gmail.com
}

\begin{document}

\maketitle

\begin{abstract}
Graph Neural Networks (GNNs) have been popularly used for analyzing non-Euclidean data such as social network data and biological data. 
Despite their success, the design of graph neural networks requires a lot of manual work and domain knowledge. In this paper, we propose a Graph Neural Architecture Search method (\textit{GraphNAS} for short) that enables automatic search of the best graph neural architecture based on reinforcement learning. Specifically, GraphNAS first uses a recurrent network to generate variable-length strings that describe the architectures of graph neural networks, and then trains the recurrent network with reinforcement learning to maximize the expected accuracy of the generated architectures on a validation data set. Extensive experimental results on node classification tasks in both transductive and inductive learning settings demonstrate that GraphNAS can achieve consistently better performance on the Cora, Citeseer, Pubmed citation network, and protein-protein interaction network. On node classification tasks, GraphNAS can design a novel network architecture that rivals the best human-invented architecture in terms of test set accuracy.
\end{abstract}

\section{Introduction}
Graph Neural Networks (GNNs) have been popularly used for analyzing graph data such as social network data and biological data. 
The basic idea of GNNs such as GraphSAGE ~\cite{GraphSAGE} is to propagate feature information between neighboring nodes so that nodes can learn feature representations by using locally connected graph structure information. 

Although GNNs have achieved great success, one shorting is to tune many parameters of the graph neural architectures. Similar to CNNs that contain many manually setting parameters such as the sizes of filters, the types of pooling layers and residual connections, tuning the parameters of GNNs generally takes heavy manual work which requires domain knowledge as well. 

Recently we observe that reinforcement learning has been successfully used to generate accurate neural architectures for CNNs and RNNs. The seminar work NAS~\cite{NAS} uses a recurrent network as controller to generate CNN and RNN network descriptions (which are referred to as child networks), and then uses the validation results of the child networks as feedback of the controller to maximize the expected accuracy of the generated architectures of the CNNs and RNNs. According to the experimental reports, the NAS search algorithm can improve CNNs and RNNs on benchmark data by percentage of 0.09 on the CIFAR-10 data and 3.6 perplexity on the Penn Treebank data. Inspired by NAS, a large body of advanced neural architecture search algorithms based on reinforcement learning have been proposed to improve its efficient and accuracy, such as the Efficient neural architecture search algorithm (ENAS ~\cite{ENAS}) and Stochastic Neural Architecture Search algorithm (SNAS) ~\cite{Xie2018SNASSN}. 

The promising results of using NAS to search neural architectures for CNNs and RNNs motivate us to use reinforcement learning to search graph neural architectures in this work. Our idea is similar to NAS for CNNs and RNNs that first uses a recurrent network as controller to generate the descriptions of GNNs and then compute the rewards of the GNNs as feedback of the controller to maximize the expected accuracy of the generated architectures of the GNNs. However, when using NAS for graph neural architecture search, the following new challenges need to be addressed:  

\begin{itemize}
	\item \textit{Challenge 1}. How to design the search space of GNNs.
	Different from CNNs for processing regular grid-structural inputs, GNNs for processing graph data that are non-Euclidean and irregularly distributed in a feature space generally contain both spatial and convolutional descriptions \cite{GraphSAGE} and GCN \cite{kipf2017semi} . 
	\item \textit{Challenge 2}. How to design an efficient reinforcement learning search algorithm. The search space of GNNs are generally very large. When generating the descriptions for GNNs by the controller, the training of reinforcement learning converges slowly.  
	\item \textit{Challenge 3}. How to evaluate the performance of the algorithm in both transductive and inductive learning settings is the third challenge. 
\end{itemize}

In this paper, we present an efficient \textit{Graph Neural Architecture Search} algorithm GraphNAS that can automatically generate neural architecture for GNNs by using reinforcement learning. To solve \textit{Challenge 1}, GraphNAS designs a search space covering sampling functions, aggregation functions and gated functions. To solve \textit{Challenge 2}, GraphNAS uses a new parameters sharing and architecture search algorithm that is more efficient than NAS for CNNs and RNNs. To solve \textit{Challenge 3}, we validate the performance of GraphNAS on node classification tasks in both transductive and inductive learning settings. The results demonstrate that GraphNAS can achieve consistently better performance on the Cora, Citeseer, Pubmed citation network, and protein-protein network. The contribution of the paper is fourfolder: 
\begin{itemize}
	\item We first study the problem of using reinforcement learning to search graph neural architectures, which has the potential to save a lot of manual work for designing graph neural architectures. 
	\item We present a new GraphNAS algorithm that can efficiently search the graph neural architectures in a large search space. 
	\item We validate the performance of GraphNAS on real-world data sets. The results show that GraphNAS can design a graph neural architecture that rivals the best human-invented architecture in terms of accuracy and F1 score on test sets.
	\item We publish the Python codes based on Pytorch for future comparisons at: 
	https://github.com/GraphNAS/GraphNAS-simple
\end{itemize}

\section{Related work}

\textbf{Graph Neural Networks}. The notation of graph neural networks was firstly outlined in the work ~\cite{Gori2005A}. Inspired by the convolutional networks in computer vision, a large number of methods that re-define the notation of convolution filter for graph data have been proposed recently. convolution filters for graph data fall into two categories, spectral-based and spatial-based. 

As spectral methods usually handle the whole graph simultaneously and are difficult to parallel or scale to large graphs, spatial-based graph convolutional networks have rapidly developed recently ~\cite{GraphSAGE,Monti2017GeometricDL,Niepert2016LearningCN,Gao2018LargeScaleLG,GAT}. These methods directly perform the convolution in the graph domain by aggregating the neighbor nodes’ information. Together with sampling strategies, the computation can be performed in a batch of nodes instead of the whole graph ~\cite{GraphSAGE,Gao2018LargeScaleLG}.

Recent graph neural architectures follow the neighborhood aggregation scheme that consists of three types of functions, i.e., neighbor sampling, correlation measurement, and information aggregation. Each layer of GNNs includes the combination of the three types of functions. For example, each layer of semi-GCN~\cite{kipf2017semi} consists of the first-order neighbor sampling, correlation measured by node's degree and the aggregate function. 

In this paper, we use reinforcement learning to search the best combination of these types of functions on each layer of GNNs, instead of manual settings in the previous work.

\noindent\textbf{Neural architecture search}.
Neural architecture search (NAS) has been popularly used to design convolutional architectures for classification tasks with image and text streaming as input~\cite{NAS,ENAS,Xie2018SNASSN,Zoph2018LearningTA,Bello2017NeuralOS}.

The seminal work of using reinforcement learning for neural architecture search aims to automatically design deep neural architecture by using a recurrent network to generate structure description of CNNs and RNNs. Following the NAS, Evolution-based NAS such as work in ~\cite{Real2017LargeScaleEO,Real2018RegularizedEF} employs evolution algorithm to simultaneously optimize topology alongside with parameters. However, evolution-based methods take enormous computational time and could not leverage the efficient gradient back-propagation. To achieve the state-of-the-art performance as human-designed architectures, the work~\cite{Real2018RegularizedEF} takes 3150 GPU days for the whole evolution. 

In comparison, the work ENAS ~\cite{ENAS} is end-to-end for gradient back-propagation.  To get rid of the architecture sampling process, DARTS ~\cite{Liu2018DARTSDA} replace the feedback triggered by constant rewards in reinforcement learning with more efficient gradient feedback from generic loss.

\section{Methods}
In this section, we first introduce the problem of searching graph neural architectures with reinforcement learning. Then, we establish the search space and we discuss an efficient search algorithm based on policy gradient descent and the parameter sharing method during training. 

\subsection{Problem formulation}
Given the search space of a graph neural architecture $\mathcal{M}$, we aim to find the best architecture $m^{*} \in \mathcal{M}$ that maximizes the accuracy $\mathcal{R}$ of the network on a validation set $D$. Here we use reinforcement learning to obtain $m^{*}$ by sampling from feasible architectures in the space $\mathcal{M}$ based on the rewards  $\mathcal{R}$ observed on $D$. %

\begin{figure}
	\includegraphics[width=0.48\textwidth]{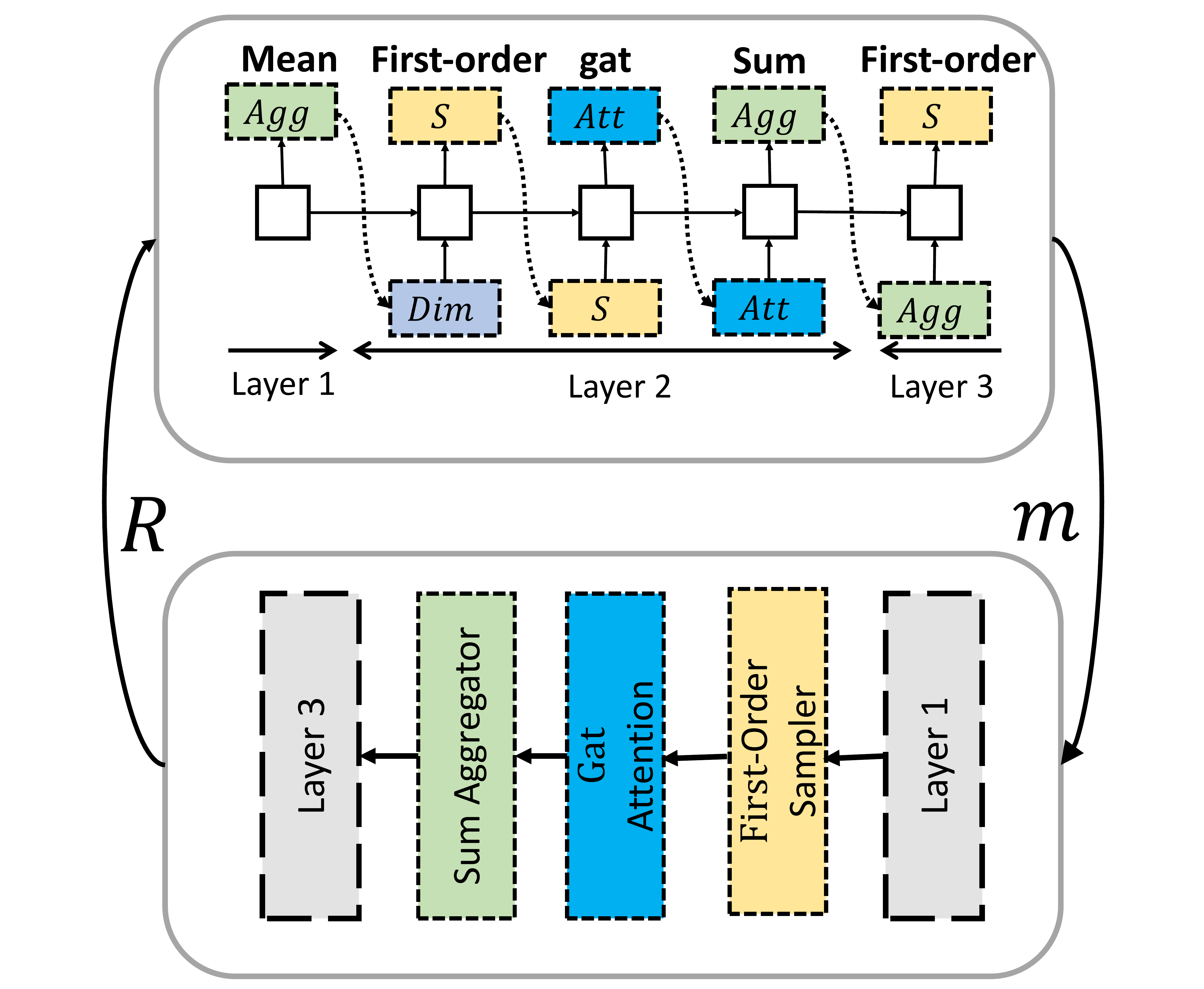}
	\caption{A simple illustration of GraphNAS. A recurrent network (the upper part) generates descriptions of graph neural architectures (the lower part), and then the validation results of the generated GNNs are used as feedback of the recurrent network (the upper part) to maximize the expected accuracy of the generated graph neural architecture (the lower part). The actions showed in current picture is not complete. All actions are described in Section~\ref{Search Space} }
	\label{fig:GraphNAS-RL}
\end{figure}

\textbf{Figure~\ref{fig:GraphNAS-RL}} shows the entire reinforcement learning framework. First, the recurrent network generates network descriptions of GNNs. Then, the generated GNNs are tested on the given validate set $D$ and the test results are used as feedback of the recurrent network. The iteration maximizes the expected accuracy of the generated GNNs on the set $D$. 

\label{formulation}

Formally, during the learning process, the recurrent network as the controller maximizes the expected accuracy $\mathcal{E}_{P(m;\theta)}[\mathcal{R}(m(w^{*},\mathcal{G}))]$ on the validation set $D$, where $P(m;\theta)$ is the distribution of $m$ parameterized by the choice of controller $\theta$, and the shared weights $w^{*}$ describing the architecture. The learning is to minimize the training loss $\mathcal{L}_{train}(m(w,\mathcal{G}))$ which can be represented as a bi-level optimization problem listed below, 

\begin{small}
	\begin{align}
	max_{w} &  ~~ \mathcal{E}[\mathcal{R}(m(w^{*},\mathcal{G})))] \nonumber \\
	s.t.: & \ w^{*}=argmin_{w} ~ \mathcal{L}_{train}(m(w,\mathcal{G})) . 
	\end{align}
\end{small}

The training process of Eq.(1) will be discussed in the remaining parts of this section. The goal of GraphNAS is to find $m^{*}$ that maximizes the expected validation accuracy $ \mathcal{E}[\mathcal{R}(m(\mathcal{G}))]$.

\subsection{Search Space}
\label{Search Space}
In GraphNAS, we use a controller network to generate the descriptions of GNNs. The controller network used in GraphNAS is implemented as a recurrent neural network which requires a state space. In order to define the search space, we introduce a generalized framework of GNNs, where each layer can be described as follows, 
\begin{table}[]
	\caption{Attention functions }
	\label{operators}
	\begin{tabular}{l|l}
		\hline
		Attention & Formula \\ \hline
		const & $e_{ij}^{con} = 1$ \\ \hline
		gcn & $e^{gcn}_{ij} = 1/\sqrt{d_{i}d_{j}}$ \\ \hline
		gat &	$e_{ij}^{gat} = leaky\_relu((W_{l} *h_{i} + W_{r} *h_{j}))$   \\ \hline
		sym-gat  &	$e_{ij}^{sym} = e_{ji}^{gat} + e_{ij}^{gat} $   \\ \hline
		cos &	$e_{ij}^{cos} =  <W_{l} *h_{i} , W_{r} *h_{j}>$   \\ \hline
		linear &	$e_{ij}^{lin} = tanh(sum (W_{l} *h_{j} ) )$   \\ \hline
		gene-linear &	$e_{ij}^{gat} = W_{a}*tanh(W_{l} *h_{i} + W_{r} *h_{j} )$   \\ \hline
		
	\end{tabular}
\end{table}
\begin{enumerate}

	\item \textbf{Feature transform functions}. The hidden embedding $H^{i-1}$ ($H^{0}$ represents the initial input) is multiplied by a weight matrix $W_{T}$, which is used to extract features and reduce feature dimensions. For each layer of GNNs, the output dimension $\mathcal{DIM}$ is required.
	
	\item \textbf{Sampling functions.} Select the receptive field $N(v)$ for a given target node. Many GNNs samples the first-order neighbors iteratively and collect messages globally. GraphSAGE and PinSAGE sample a fixed size neighbor to speed up for large graphs. FastGCN uses importance sampling while maintaining the performance of the algorithm. LGCN sorts the first-order neighbors' features and  selects top-$k$ features.
	For each layer, one of the sampling methods $\mathcal{SAM}$ is required.
	
	\item \textbf{Correlation measure functions.} Calculate the correlation of node $v$ with its neighbors $N(v)$. GAT assigns neighborhood importance by using attention layers. Semi-GCN assigns neighborhood importance according to the degree of nodes. More choices are listed in Table \ref{operators}.
	For each layer, we choose one correlation measurement method $\mathcal{ATT}$ and repeat times $\mathcal{K}$.
	
	\item \textbf{Aggregation functions.}  Aggregate data from neighbors to generate an embedding for each node $v$. Most GNNs use the sum aggregator, Mean aggregator, LSTM aggregator and pooling aggregator. MLP aggregator are described in the work~\cite{Xu2018HowPA}. 
	For each layer, one of the aggregation functions $\mathcal{AGG}$ is required.
	
	\item \textbf{Residual functions.} Merge historical hidden representation $H^{his}$ as a part of the current embedding after a transform function. The merge function includes concatenation and adding.
	For each layer of GNNs, a previous layer's index $\mathcal{I}$ and the activation function $\mathcal{ACT}$ of the current layer are required to build a residual layer.
	
	\item \textbf{Gated functions.}  As in GeniePath~\cite{GeniePath}, the attention procedure learns the importance of neighbors with different sizes. Gated procedure extracts and filters signals aggregated from neighbors of distant hops.
\end{enumerate}

To sum up, we define the search space of GNNs $\mathcal{M}$ as follows: the sampling dimension $\mathcal{SAM}$ , the correlation measure dimension $\mathcal{ATT}$, the aggregation dimension $\mathcal{AGG}$, the numbers of multi-heads $\mathcal{K}$, the output hidden embedding $\mathcal{DIM}$, the previous layers' index $\mathcal{I}$ and the activation function $\mathcal{ACT}$. As a result,  GraphNAS can generate the architecture descriptions as a sequence of tokens. Each token represents one of the functions in the architecture space $\mathcal{M}$. 

Note that we do not predict training parameters such as the learning rates of GNNs and we also assume that the architectures without gated procedures bring large computation and improvement during the initial stage of training. It is possible to add those actions as one of the predictions. In our experiments, the process of generating an architecture stops if the number of layers exceeds a preset value.

\begin{table}
	\caption{Action operators in search space}
	\label{actions}
	\begin{tabular}{l|l}
		\hline
		
		action            & operators  
        \\ \hline
        sample method & "first-order"    \\ \hline
		attention type    & listed in Table \ref{operators}      \\ \hline
		aggregation type & "sum", "mean-pooling", "max pooling", \\ & "mlp" \\  \hline
		activation type  & "sigmoid", "tanh", "relu", "linear",\\ &"softplus", "leaky\_relu", "relu6", "elu" \\ \hline
		number of heads & 1,2,4,6,8,16                                                                   \\ \hline
		hidden units     & 4, 8, 16, 32, 64, 128,256                                                         \\ \hline
	\end{tabular}
\end{table}

\subsection{Search Algorithm}
\label{Algorithms}
\textbf{Training the controller parameters $\theta$.}
In order to maximize the objective function given in Eq.(1), we describe a policy gradient method to update parameters $\theta$ so that the controller network generates better architectures over time.

The architecture descriptions (hyperparameters) of GNNs that the controller predicts can be viewed as a list of actions $m_{1:T}$. GNNs will achieve an accuracy of $m(w,\mathcal{G})$ on a held-out dataset at convergence. We use the accuracy $\mathcal{R}$ as reward signal and use reinforcement learning to train the controller. Since the reward signal $\mathcal{R}$ is non-differentiable, we use a policy gradient method to iteratively update $\theta$ with a moving average baseline for reward to reduce variance~\cite{Williams1992Simple} as follows, 
\begin{small}
	\begin{eqnarray}
	& &\nabla_{\theta}\mathcal{E}_{{P(m_{1:T};\theta)}}[R(m(w,\mathcal{G}))]~\\ \cdot
	& & =\sum_{t=1}^{T}\mathcal{E}_{P(m_{1:T};\theta)}[\nabla_{\theta}logP(m_{t}|m_{t-1:1};\theta)R(m(w,\mathcal{G}))]	\nonumber
	\end{eqnarray}
\end{small}

\textbf{Training the shared parameters $w $.}
In this step, we use stochastic gradient descent (SGD) with respect to $w$ to minimize the training loss $\mathcal{L}_{train}$, where  $\mathcal{L}_{train}$  is the standard cross-entropy loss in node classification, obtained from a minibatch of training data with a model $m$ sampled by the controller.

\subsection{Parameters Sharing}
In most previous neural architecture search algorithms, generated models are trained from the scratch. However, training  models from scratch to convergence brings heavy computation. Recently, ENAS~\cite{ENAS} forces all child models to share weights in order to improve the efficiency.  Similarly, we introduce parameters sharing for GraphNAS.

Parameters sharing in ENAS are without conditions, ignoring the performance of the child models. This strategy does not suit for GNNs. As we found in experiments, parameters sharing between different GNNs does not work immediately, but can be observed after several iterations. Therefore we use a new strategy to share weights between different GNNs. 

\begin{figure}
	\includegraphics[width=0.48\textwidth]{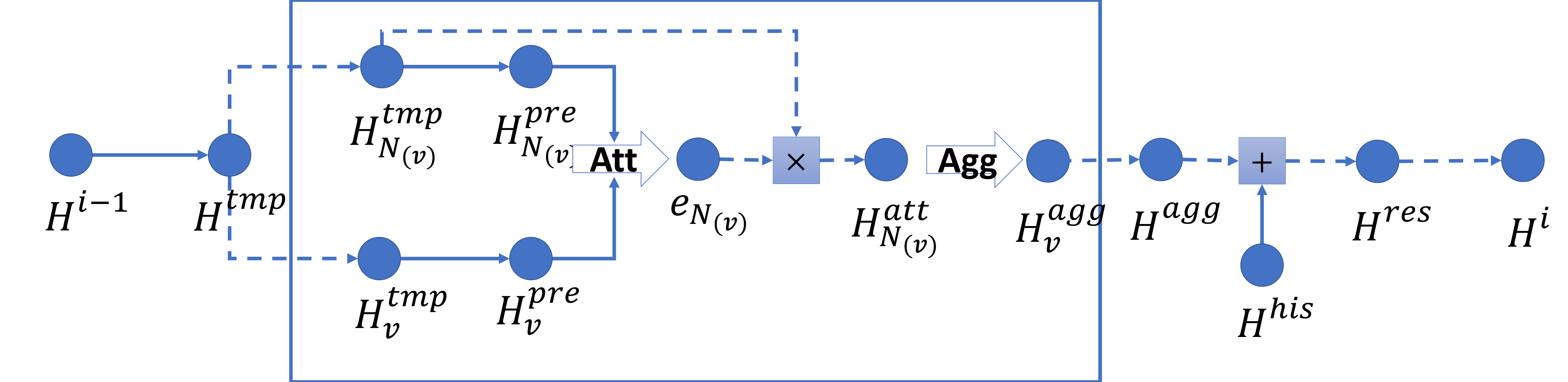}
	\caption{Parameters of each layer in GNNs. Circles represent hidden embeddings in each layer. Solid arrows represent the transform operation with parameters, dotted arrows indicate operations with no parameters, and hollow arrows with text represent the remaining functions such as Att for $\mathcal{ATT}$, Agg for $\mathcal{AGG}$).}
	\label{fig:parametersharing}
\end{figure}

\textbf{Sharing strategy.} Figure~\ref{fig:parametersharing} shows the parameters from one layer. Solid arrows represent the transform weights in GNNs which are shared between different GNNs including $W_{T}$, $W_{l}$, $W_{l}$ and $W_{res}$. However, GNNs constraints search dimensions such as the attention and aggregation dimensions. For instance, $W_{l}$, $W_{r}$ listed in Table \ref{operators} are used to form correlation measures for the  attention functions. 
In GraphNAS, we allow different GNNs sharing the transform weights. Parameters are only shared for specific combinations of attention and aggregation functions

\textbf{Update strategy.}  The parameters $w$ are trained and updated during training child models. The parameter update is also different from ENAS.  Parameters shared by GNNs stored in the form of dictionary. When training, GNNs obtain a copy of shared parameters. After training, the parameters of the current child model are merged into the shared parameters $w$ when the reward is positive.



\label{Parameters share}


\textbf{Reward generation.}
In ENAS \cite{ENAS}, the reward is generated according to the shared parameters without training child models.
Since there may be parameters untrained, the strategy may cause deviation during reward generation. In GraphNAS, we train the child models with shared parameters to obtain more precise reward. After then, we apply a moving average on rewards to generate the final reward.

\textbf{Exploration for shared parameters.}
Models with updated sharing parameters generally have large reward. Therefore, the controller has the potential to choose structures appearing at the beginning.
In order to avoid this bias, we allow the GraphNAS to do exploration. During exploration, the parameters $\theta$ of the controller are fixed, while the shared parameters $w$ are trained with novel structures.

\textbf{Deriving architectures.}
We derive novel architectures from a trained GraphNAS model. We first sample several models under the distribution of $P(m,\theta)$. For each sampled model, we compute its reward on a single minibatch sampled from the validation set after a few iterations. We then take only the model with the highest reward to re-train from scratch. It is possible to improve the results by training all the sampled models from scratch and select the model with the highest accuracy on a separated validation set~\cite{NAS,Zoph2018LearningTA}.


\section{Experiments}
We test the performance of GraphNAS on both transductive and inductive learning scenarios. We use citation networks including Cora, Citeseer, and Pubmed for the transductive learning, and PPI for the inductive learning. On each dataset, we have a separated held-out validation set used to generate reward to compute the reward signal. The reported performance on the test set is computed only once for the network that achieves the best result on the held-out validation set.

\subsection{Datasets}
\textbf{Transductive Learning Task}
We classify academic papers into different subjects using the Cora, Citeseer and Pubmed datasets. The data obtained from semi-GCN \cite{kipf2017semi} has been preprocessed. We follow the same setting used in semi-GCN that allows 20 nodes per class for training, 500 nodes for validation and 1,000 nodes for testing. 

\textbf{Inductive Learning Task}
 We use the protein-protein interaction (PPI) dataset, which contains 20 graphs for training, two graphs for validation, and two graphs for testing. Since the graphs for validation and testing are separated, the training process does not use them. There are 2,372 nodes in each graph on average. Each node has 50 features including positional, motif genes and signatures. Each node has multiple labels from 121 classes.

\subsection{Baseline Methods}
In order to evaluate the GNNs searched by GraphNAS, we choose the state-of-the-art GNNs for comparisons, 
\begin{itemize}
	\item Chebyshev ~\cite{Defferrard2016ConvolutionalNN}. The method approximates the graph spectral convolutions by a truncated expansion in terms of Chebyshev polynomials up to T-th order. This method needs the graph Laplacian in advance, so that it only works in the transductive setting.
	\item Semi-GCN ~\cite{kipf2017semi} is the same as Chebyshev, it works only in the transductive setting. 
	\item GraphSAGE ~\cite{GraphSAGE}  consists of a group of inductive graph representations with different aggregation functions. The GCN-mean with residual connections is equivalent to GraphSAGE using mean pooling.
	\item GAT~\cite{GAT} introduces attention into GNN. Therefore, GAT  archives good results in both transductive and inductive learning.
	\item LGCN ~\cite{Gao2018LargeScaleLG}  enables regular convolutional operations on generic graphs which archives good results in both transductive and inductive learning.
\end{itemize}

All the tasks in transductive learning are single-label classification. We use accuracy as the measure for comparison. On the other hand, tasks in transductive learning are multi-label classification, and we use F1 score as the measure.

\subsection{Architecture on Transductive learning}

\textbf{Search space.} Our search space consists of the functions listed in Section~\ref{Search Space}. For each layer of GNNs, the controller  has to sample actions $m \in {\mathcal{DIM},\mathcal{SAM},\mathcal{ATT},\mathcal{K},\mathcal{AGG},\mathcal{ACT}}$ which do not contain previous layer index $\mathcal{I}$ for skip connection. In experiments on the citation dataset, there are usually two layers for GNNs. 

\textbf{Training details.} The controller is a one-layer LSTM with 100 hidden units. It is trained with the ADAM optimizer ~\cite{Kingma2015AdamAM} with a learning rate of 0.0035. The weights of the controller are initialized uniformly between -0.1 and 0.1. To prevent premature convergence, we also use a tanh of 2.5 and a temperature of 5.0 for the sampling logits ~\cite{Bello2017NeuralOS,Bello2016NeuralCO},
and add the controller’s sample entropy to the reward, weighted by 0.0001. 

Once the controller samples an architecture, a child model is constructed and trained for 200 epochs without parameter sharing. During training, we apply L2 regularization with $\lambda$  = 0.0005 for Cora and Citeseer. Furthermore, dropout with $p$ = 0.6 is applied to both layers’ inputs, as well as to the normalized attention coefficients. For Pubmed, we set L2 regularization to $\lambda$  = 0.001. 

In all experiments, child models are initialized using the Glorot initialization ~\cite{Glorot2010UnderstandingTD} and trained to minimize cross-entropy loss on the training nodes using the Adam SGD optimizer ~\cite{Kingma2015AdamAM} with an initial learning rate of 0.01 for Pubmed, and 0.005 for all the other datasets. In both cases we use an early stopping strategy according to  the cross-entropy loss and accuracy on the validation nodes, with a patience of 100 epochs.

During training the controller, we fix the number of child network layers to be two, because many GNNs obtain the best performance on these dataset with two layers. 
Besides, we do not force GNNs sharing parameters, because training GNNs to convergence are fast on these datasets and models are easy to over-fitting in a semi-supervision task.

\textbf{Results.} After the controller trains 1,000 architectures, we collect the top 5 architectures that achieve the best validation accuracy. Then, we compute the test accuracy and time for each epoch of such models and summarize their results in Table ~\ref{table:cora}. The time reported here is the training time for running one epoch using a single 1080Ti GPU. As can be seen from the table, GraphNAS can design several promising architectures that perform as well as some of the best models on this dataset.
Other experiments results on citation network are listed in Table ~\ref{other_citation}. 

\begin{figure}
	\includegraphics*[width=.45\textwidth, height=150pt]{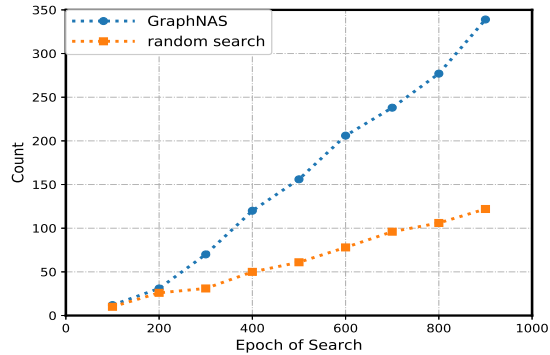}
	\caption{The number of GNNs whose accuracy over 0.81 on the validation set during search. Red line stands for GraphNAS, blue line stands for random search. GraphNAS outperforms random search. }
	\label{fig:random}
\end{figure}

\textbf{Random Search.} Besides reinforcement learning, one can use random search to find the network. Although this baseline seems simple, it is often very hard to surpass. We report the number of GNNs model which has accuracy over 0.81 on validation set during search in Figure ~\ref{fig:random}. And we list the best structure found by random search in Table\ref{table:cora}. The results show that GraphNAS trends to find better GNNs.
\begin{table}[]
	\caption{Performance of GraphNAS and the state-of-the-art on Cora.}
	\label{table:cora}
	\begin{tabular}{l|ll|ll}
		\hline
		Models    & Depth & Params &  Time(s) & Accuracy   \\ \hline
		Chebyshev &   2    &      92K   & 0.49   & 81.2\%       \\
		GCN      &    2   &     23K & 0.08  & 81.5\%       \\
		GAT       &    2   &    237K & 0.62 & 83.0$\pm$0.7\% \\
		LGCN      &  2      &    56K        & 0.14 & 83.3$\pm$0.5\% \\ \hline
		random & 2  &        364K          &	1.29		&82.0$\pm$0.6\%				\\
		GraphNAS   &  2   &  188K         &   0.13 &  84.2$\pm$1.0\%         \\ \hline
		
	\end{tabular}
\end{table}

\begin{table}[]
	\caption{Performance of GraphNAS and the state-of-the-art models on Citeseer and Pubmed in term of accuracy.}
	\label{other_citation}
	\begin{tabular}{l|l|l}
		\hline
		Models    &    Citeseer     & Pubmed       \\ \hline
		Chebyshev &        69.8\%       & 74.4\%       \\
		GCN       &  70.3\%       & 79.0\%       \\
		GAT       &     72.5$\pm$0.7\%         &      79.0$\pm$0.3\%        \\
		LGCN      & 73.0$\pm$0.6\% & 79.5$\pm$0.2\% \\ \hline
		GraphNAS  &      73.1$\pm$0.9\%     & 79.6$\pm$0.4\%      \\ \hline
	\end{tabular}
\end{table}

\subsection{Architecture on Inductive Learning}

\textbf{Search space.} 
We use the full search space defined in Section~\ref{Search Space} . For skip-connection, we perform two sets of experiments, where one fixes the input of residual layer with output of the last hidden layer and the other allows the controller to predict previous layer index to build skip-connection.

\textbf{Training details.}
The setting of training  and the controller  are the same as  in transductive learning.
We use parameter sharing to solve the huge computational resource requirements.
The shared parameters of the child models are trained using the Adam SGD  optimizer with a learning rate of 0.005. 
Before training the controller, the exploration process is executed at the first 20 epochs. After that the controller is trained for 50 epochs.
During training of the child model, we apply no L2 regularization and dropout. During the  process,   each GNN  model  sampled by the controller is trained for five epochs with shared parameters. 

During the training of the controller, we fix the number of child networks layers at three, because most  GNNs obtain the best performance on this dataset are with three layers.

\textbf{Results.}
After the controller trained for 1,000 time, we let the controller to output the best model from 200 sampled GNNs. And we then compute the  micro-f1 score and the time for each epoch of the model and summarize the results in Table \ref{PPI result}. The time reported here is the training time for running one epoch using a single 1080Ti GPU. As can be seen from the table, GraphNAS can design several promising architectures that perform as well as the best models on this dataset.

\begin{figure}
	\includegraphics*[width=.48\textwidth, height=170pt]{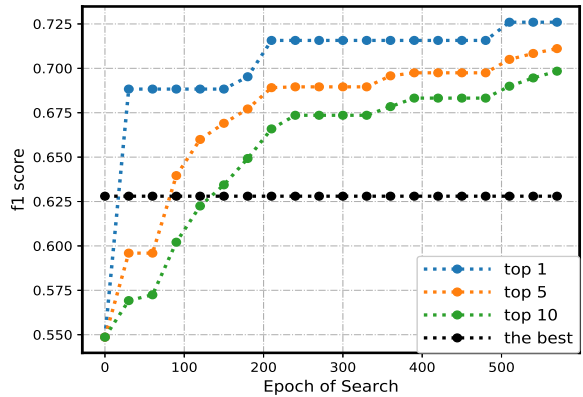}
	\caption{F1 score of the best GNNs on the ppi validation set during GraphNAS search. Blue is the best, Green is the average of top 5, Red is the average of top 10, Black is the best search in 600 structures which appears at the search $500^{th}$.    }
	\label{fig:runing_process}
\end{figure}

\begin{table}[]
	\caption{Performance of GraphNAS and the state-of-the-art on PPI.}
	\label{PPI result}
	\begin{tabular}{l|ll|l}
		\hline
		Models    & Depth & Params  & micro-F1(\%)     \\ \hline
		GraphSAGE(lstm) &  2     &  0.26M       & 61.2       \\
		GeniePath & 3  & 1.81M   &97.9  \\
		GAT       &   3    &  3.64M      & 97.3$\pm$0.2 \\
		LGCN      &  -     & -     & 77.2$\pm$0.2 \\ \hline

		GraphNAS( no sc) &	3	&	3.95M	&	 98.6$\pm$0.1			\\

		GraphNAS with sc&	3	&	2.11M	&	 	  97.7$\pm$0.2 \\
 
		NAS-like search &  3     &  0.95M    &  95.7$\pm$0.2                 \\ 
 
		ENAS-like search&  3     &  1.38M     &   96.5$\pm$0.2             \\ \hline
	\end{tabular}
\end{table}

\textbf{Effectiveness of parameters sharing.} To evaluate the effectiveness of parameter sharing strategy during search, we compare the F1 score of the best structure designed by GraphNAS with parameters sharing and trained from scratch in Figure ~\ref{fig:runing_process}. Both of them are trained for five epochs. 

\begin{figure}
	\includegraphics*[width=.48\textwidth, height=170pt]{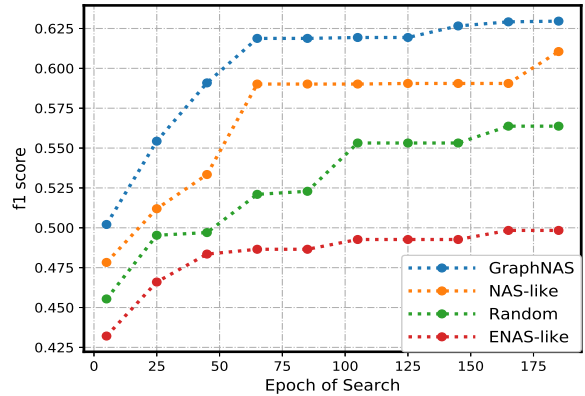}
	\caption{GraphNAS compares with \textit{random search}, \textit{NAS-like GraphNAS}, and \textit{ ENAS-like GraphNAS} on ppi dataset. GrpahNAS has the best F1 score.}
	\label{fig:ppi_compare_random}
\end{figure}

\textbf{Comparison against Search strategy.} We compare GraphNAS with other search strategies including random search, reinforcement learning without parameter sharing (NAS-like), and GraphNAS without iterations of GNNs during training the controller (ENAS-like). Each sampled GNN is trained for two epochs.
We show the model's validation F1-score during training in Figure ~\ref{fig:ppi_compare_random}. The performance of the found best model are listed in Table\ref{PPI result}. 
The results show that GraphNAS trends to find a better GNN model.

\section{Conclusions}
In this paper, we study a new problem of graph neural architecture search with reinforcement learning. We present a GraphNAS algorithm that can design accurate graph neural network architectures that rival the best human-invented architectures in terms of test set accuracy. Experiments on node classification tasks in both transductive and inductive learning settings demonstrate that GraphNAS can achieve consistently better performance on citation networks, and protein-protein interaction network. Comparisons with existing search strategies show that the new parameters sharing and search strategy used in GraphNAS are effective.
\newpage 
\newpage

\appendix

\bibliographystyle{named}
\bibliography{GraphNAS}

\end{document}